\documentclass[letterpaper, 10 pt, journal, twoside]{IEEEtran} 

\usepackage[top=60pt,bottom=55pt,rmargin=48.5pt,lmargin=48.5pt]{geometry}

\usepackage{graphicx}
\usepackage{algorithm}
\usepackage{algpseudocode}

\usepackage{amsmath}
\usepackage{mathtools}
\usepackage{amsfonts}
\usepackage{verbatim}
\usepackage{float}
\usepackage{hyperref}
\usepackage{subcaption}

\newcommand{\Tr}{\mathop{\bf Tr}}

\usepackage{pgfplots}
\pgfplotsset{compat=1.15,
	legend style={font=\footnotesize},
}
\usepackage{tikzscale}
\usetikzlibrary{matrix}

\usetikzlibrary{arrows.meta}

\usetikzlibrary{backgrounds}
\usepackage{pgfplots}
\usepgfplotslibrary{groupplots}

\usepackage{booktabs}
\usepackage{array}

\newcommand{\specialcellbold}[2][c]{%
	\bfseries
	\begin{tabular}[#1]{@{}l@{}}#2\end{tabular}%
}
\newcolumntype{P}[1]{>{\centering\arraybackslash}p{#1}}

\newif\ifproofread

\newcommand{\rev}[1]{%
	\ifproofread
	\textcolor{red}{#1}%
	\else
	\textcolor{black}{#1}%
	\fi
}

\proofreadfalse

\definecolor{orange}{RGB}{255,127,0}
\definecolor{cyan}{RGB}{0,255,255}
\definecolor{magenta}{RGB}{255,0,255}

\begin{document}

\title{Direct Policy Optimization\\using Deterministic Sampling and Collocation}

\author{Taylor A. Howell$^{1}$, Chunjiang Fu$^{2}$, and Zachary Manchester$^{3}$% <-this % stops a space
	\thanks{$^{1}$Taylor A. Howell is with the Department of Mechanical Engineering, Stanford University,
		Stanford, CA 94305, USA
		{\tt\footnotesize thowell@stanford.edu}}%
	\thanks{$^{2}$Chunjiang Fu is with Frontier Robotics, Innovative Research Excellence, Honda R\&D Co., Ltd, Wako-shi, Saitama 3510188, Japan 
		{\tt\footnotesize chunjiang\_fu@jp.honda}}%
	\thanks{$^{3}$Zachary Manchester is with The Robotics Institute, Carnegie Mellon University,
		Pittsburgh, PA 15213, USA
		{\tt\footnotesize zacm@cmu.edu}}%
}

%\markboth{IEEE Robotics and Automation Letters. Preprint Version. February, 2021}
%{Howell \MakeLowercase{\textit{et al.}}: Direct Policy Optimization using Deterministic Sampling and Collocation} 

\maketitle

\begin{abstract}
	We present an approach for approximately solving discrete-time stochastic optimal-control problems by combining direct trajectory optimization, deterministic sampling, and policy optimization. Our feedback motion-planning algorithm uses a quasi-Newton method to simultaneously optimize a \rev{reference} trajectory, a set of deterministically chosen sample trajectories, and a parameterized policy. We demonstrate that this approach exactly recovers LQR policies in the case of linear dynamics, quadratic objective, and Gaussian disturbances. We also demonstrate the algorithm on several nonlinear, underactuated robotic systems to highlight its performance and ability to handle control limits, safely avoid obstacles, and generate robust plans in the presence of unmodeled dynamics.
\end{abstract}


\section{Introduction}	
	\IEEEPARstart{T}{rajectory} optimization (TO) is a powerful tool for solving deterministic optimal-control problems in which accurate models of the system and its environment are available. However, when disturbances or unmodeled dynamics are significant, a stochastic optimal-control approach, in which a feedback policy is optimized directly, can often produce more robust performance~\cite{bertsekas2004stochastic}.
	
	Unfortunately, general solution methods for solving stochastic optimal-control problems suffer from the curse of dimensionality and are only applicable to low-dimensional systems in practice. To scale to many interesting robotic systems, approximations must be made. Typically, these include simplifying or linearizing dynamics, approximating value functions or policies with polynomials or neural networks, or approximating distributions with Gaussians or Monte Carlo sampling.
	\begin{figure}[t]
		\centering
		\includegraphics[width=.45\textwidth]{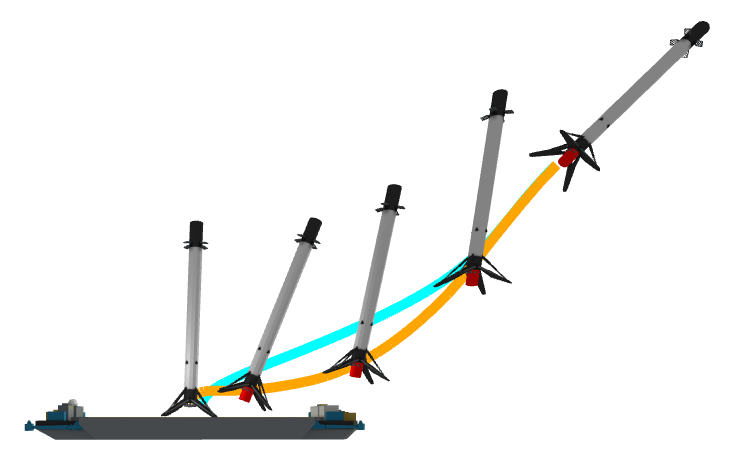}
		\caption{Rocket soft-landing position trajectories generated by TO (blue) and DPO (orange). Unlike the TO solution tracked with LQR, the DPO policy successfully lands the rocket despite fuel slosh.}
		\label{rocket_trajectory}
	\end{figure}

	We present Direct Policy Optimization (DPO), a computationally tractable algorithm for finding approximate solutions to stochastic optimal-control problems that jointly optimizes a small number of trajectories and a policy. This algorithm combines several key ideas:
	\rev{
	\begin{itemize}
		\item Joint optimization of parameterized policies along with trajectories.
		\item Deterministic sampling of trajectories and approximation of expectations using the unscented transform.
		\item Direct collocation for enforcing dynamics along trajectories without performing rollouts.
		\item Use of large-scale constrained nonlinear programming solvers based on quasi-Newton methods for fast and robust convergence.
	\end{itemize}
	}
	In contrast to many other approaches, DPO is able to easily enforce constraints like torque limits and obstacle avoidance, makes extensive use of analytical models and their derivatives, and is extremely sample efficient.
	
	We first provide background for the discrete-time stochastic optimal-control problem, present related work on feedback motion planning, and give an overview of the unscented transform in Section \ref{background}. In Section \ref{DPO}, we present the DPO algorithm. We then demonstrate that DPO exactly recovers LQR policies when the dynamics are linear, the objective is quadratic, and disturbance inputs are Gaussian, and provide examples using DPO for several nonlinear, underactuated-control problems in Section \ref{examples}. Section \ref{discussion} offers discussion of the experimental results and, finally, we summarize our work and propose directions for future research in Section \ref{conclusion}.
	
\section{Background}\label{background}
	This section provides brief reviews of the discrete-time stochastic optimal-control problem and the unscented transform, as well as a survey of related solution approaches.

\subsection{Discrete-Time Stochastic Optimal Control}
	We formulate the discrete-time stochastic optimal-control problem as:
	\begin{align}
		& \underset{\Theta}{\mbox{minimize}} && \mathbf{E} [ J(\tau) ] \label{policy_opt}\\
		& \mbox{subject to} && x_{t+1} = f_t(x_t,u_t, w_t),&& t = 1,\dots,T-1, \notag\\
		& && u_t = \pi_t(x_t, \theta_t), && t = 1,\dots,T-1, \notag\\
		& && \mathbf{Prob} (c_j(\tau) > 0 ) \leq \epsilon_j, && j = 1,\dots,k. \notag
	\end{align}
	\rev{The system's state, $x_t \in \mathbf{R}^{n}$, and control inputs, $u_t \in \mathbf{R}^{m}$, define a trajectory, $\tau = (x_1,\dots,x_T,u_1,\dots,u_{T-1}) \in \mathbf{R}^z$, with a subscript denoting the time index $t$, over a planning horizon $T$. The initial state, $x_1$, is a random variable. The discrete-time stochastic dynamics, $f_t : \mathbf{R}^n \times \mathbf{R}^m \times \mathcal{W} \rightarrow \mathbf{R}^n$, are subject to random disturbance inputs, $w_t \in \mathcal{W}$. We seek a policy, $\pi_t : \mathbf{R}^n \times \mathbf{R}^p \rightarrow \mathbf{R}^m$, parameterized by $\Theta = (\theta_1, \dots, \theta_{T-1}) \in \mathbf{R}^{p(T-1)}$, to minimize the objective, $J : \mathbf{R}^z \rightarrow \mathbf{R}$, with the expectation taken over the initial state and disturbance inputs. Chance constraints, with $c_j : \mathbf{R}^z \rightarrow \mathbf{R}$ and probability of violation less than tolerance $\epsilon_j \in \mathbf{R_{+}}$, can further constrain trajectories.}
	
\subsection{Related Work}
	Model-based approaches often tackle problem \eqref{policy_opt} in a decoupled fashion: First, generate a \rev{reference} trajectory assuming no disturbances; then, design a tracking feedback policy to reject disturbances. Using collocation methods~\cite{stryk1993numerical} or differential dynamic programming (DDP)~\cite{jacobson1970differential} to optimize a trajectory, then tracking it with a time-varying LQR controller has achieved impressive results for complex, real-world systems~\cite{kuindersma2016optimization,moore2014robust}.
	
	There are two primary drawbacks to these classic synthesis techniques: First, there is no explicit consideration of uncertainty or disturbances. Robustness is often achieved heuristically via \textit{post hoc} Monte Carlo testing and tuning. Second, by decoupling the synthesis of the \rev{reference} trajectory and policy, performance is often sacrificed.
	
	DIRTREL~\cite{manchester2019robust} optimizes a \rev{reference} trajectory with an additional cost term that penalizes a linearized approximation of the tracking error from an LQR policy. Chance constraints are approximated by enforcing constraints at a finite number of samples. There is also a variation of DDP that can account for multiplicative noise applied to the controls~\cite{todorov2005generalized}. Unlike these methods, DPO directly optimizes policies and \rev{can} propagate uncertainty through nonlinear dynamics.
	
	Several learning-based approaches exist that directly optimize feedback policies. \rev{Policy gradient methods~\cite{silver2014deterministic, williams1992simple}} use first-order or derivative-free methods to optimize parameterized policies, typically without direct access to the underlying dynamics model. Domain randomization~\cite{tobin2017domain} can be employed to vary model and environment parameters to encourage policy robustness. Random search~\cite{mania2018simple}, stochastic gradient descent~\cite{bu2019lqr}, and Newton methods~\cite{wytock2013fast} have also been used to optimize linear feedback and MPC policies~\cite{agrawal2020learning, amos2018differentiable}. A major benefit of stochastic methods is their inherent exploration of the policy space.
	
	Guided Policy Search (GPS)~\cite{levine2013guided} is a hybrid approach that alternates between optimizing sample trajectories and fitting a policy. DDP is used to generate high-reward trajectories around the current policy that are subsequently employed to improve the policy. This procedure is then iterated with a regularizer to keep new trajectories close to those generated by the previous policy.
		
	While GPS does not make strong assumptions about state and disturbance distributions or explicitly require dynamics models, it relies heavily on Monte Carlo techniques that require a large number of samples. As a result, training can require significant time and computational resources. In contrast, DPO leverages analytical models and uses only a small number of deterministically chosen samples, making it much more efficient.
	
	Finally, we note that many of the approaches outlined in this section---including DDP and many of the policy gradient methods---rely on explicit forward simulation or ``rollouts'' along with some form of backpropagation. These techniques are vulnerable to numerical ill-conditioning issues colloquially known as the ``tail-wagging-the-dog problem'' in control and the ``vanishing'' or ``exploding'' gradient problem in reinforcement learning. In contrast, DPO employs collocation methods that simultaneously optimize state and control trajectories with dynamics enforced as constraints. Such ``direct'' methods enjoy far better numerical conditioning and robustness, especially with long-horizon plans.

\subsection{Unscented Transform} \label{ut}
	\begin{figure}[t]
		\begin{center}
			\includegraphics[height=3.5cm]{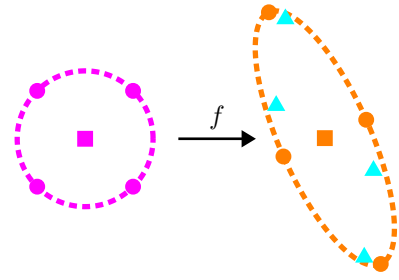}
		\end{center}
		\caption{Unscented transform visualized in 2D for initial (left) and transformed (right) distributions. Sigma points (circles) with sample mean (square) and sample covariance (dashed) are propagated through a nonlinear function (\rev{triangles}) and then resampled to compute new sigma points.}
		\label{unscented}
	\end{figure}
	The unscented transform is a procedure for propagating a unimodal probability distribution through a nonlinear function using deterministic samples, often referred to as sigma points. This tool is commonly used for state estimation, where it is generally considered to be superior to the extended Kalman filter's linear propagation of covariance matrices~\cite{uhlmann1995dynamic}. There are many variations of the unscented transform~\cite{menegaz2015systematization}; the version we utilize is visualized in Fig. \ref{unscented} and outlined below.
	
	Assuming a unimodal state distribution with mean $\mu_t \in \mathbf{R}^n$, covariance $P_t \in \mathbf{S}_{++}^{n}$, and an uncorrelated zero-mean disturbance distribution with covariance $D_t \in \mathbf{S}_{++}^{d}$, we generate $N = 2(n + d)$ sigma points:
	\begin{equation}
		\begin{bmatrix} x_t^{(i)} \\ w_t^{(i)} \end{bmatrix} \leftarrow \begin{bmatrix} \mu_t \\ 0 \end{bmatrix} \pm \beta_t \, \mathbf{col} \Bigg( \sqrt{\begin{bmatrix}P_t & 0 \\ 0 & D_t \end{bmatrix}} \Bigg).
		\label{sigma_point_update}
	\end{equation}
	These samples, denoted with superscripts, are constructed using the column vectors of a square root of the joint covariance matrix. In this work we use the principle (symmetric) matrix square root, although other decompositions, such as the Cholesky factorization, could be employed. The parameter $\beta_t \in \mathbf{R}_{+}$ \rev{controls} the spread of the samples around the mean. \rev{For a Gaussian distribution propagated through linear dynamics, the unscented transform will exactly recover the updated distribution. For nonlinear systems, the selection of $\beta$ can affect performance, and has been explored extensively in the literature on unscented Kalman filters~\cite{menegaz2015systematization}.}
	
	\begin{algorithm}[t]
		\caption{\rev{Unscented} dynamics}\label{unscented_dynamics}
		\begin{algorithmic}[1]
			\Function{$g_t$}{$\tau_t^{(1:N)}, D_t$}
			\For{$i = 1:N$}
			\State $x_{t}^{(i)},~w_t^{(i)} \leftarrow$ compute sigma points ($\ref{sigma_point_update}$)
			\State $u^{(i)}_t,~x^{(i)}_{t+1} \leftarrow$ propagate sigma points ($\ref{sample_propagate_control}$, $\ref{sample_propagate}$)
			\EndFor
			\State $\mu_{t+1},~\nu_t \leftarrow$ compute sample means ($\ref{sample_mean}$, $\ref{sample_mean_control}$)
			\State $P_{t+1},~L_t \leftarrow$ compute sample covariances ($\ref{sample_covariance}$, $\ref{sample_covariance_control}$)
			\State \textbf{return} $x_{t+1}^{(1:N)}$
			\EndFunction
		\end{algorithmic}
	\end{algorithm}	

	Sample points are first propagated through the policy:
	\begin{equation}
		u_t^{(i)} = \pi_t(x_t^{(i)}), \label{sample_propagate_control}
	\end{equation}
	and then sample dynamics:
	\begin{equation}
		x_{t+1}^{(i)} = f_t^{(i)}(x_t^{(i)}, u_t^{(i)}, w_t^{(i)}), \label{sample_propagate}
	\end{equation}
	in order to compute a sample mean:
	\begin{equation}
		\mu_{t+1} = \frac{1}{N} \sum\limits_{i=1}^{N} x_{t+1}^{(i)}, \label{sample_mean}
	\end{equation}
	and sample covariance:
	\begin{equation}
		\rev{P_{t+1} = \frac{1}{2 \beta_t^2}\sum\limits_{i=1}^{N} (x_{t+1}^{(i)} - \mu_{t+1})(x_{t+1}^{(i)} - \mu_{t+1})^T, \label{sample_covariance}}
	\end{equation}
	for the state distribution at the next time step. Similarly, we compute a sample mean:
	\begin{equation}
		\nu_t = \frac{1}{N} \sum\limits_{i=1}^{N} u_t^{(i)}, \label{sample_mean_control}
	\end{equation}
	and sample covariance:
	\begin{equation}
		\rev{L_t = \frac{1}{2 \beta_t^2}\sum\limits_{i=1}^{N} (u_t^{(i)} - \nu_t)(u_t^{(i)} - \nu_t)^T, \label{sample_covariance_control}}
	\end{equation}
	for the control inputs at the current time step.

	\begin{figure}[t]
		\centering
		\includegraphics[width=0.1425\textwidth]{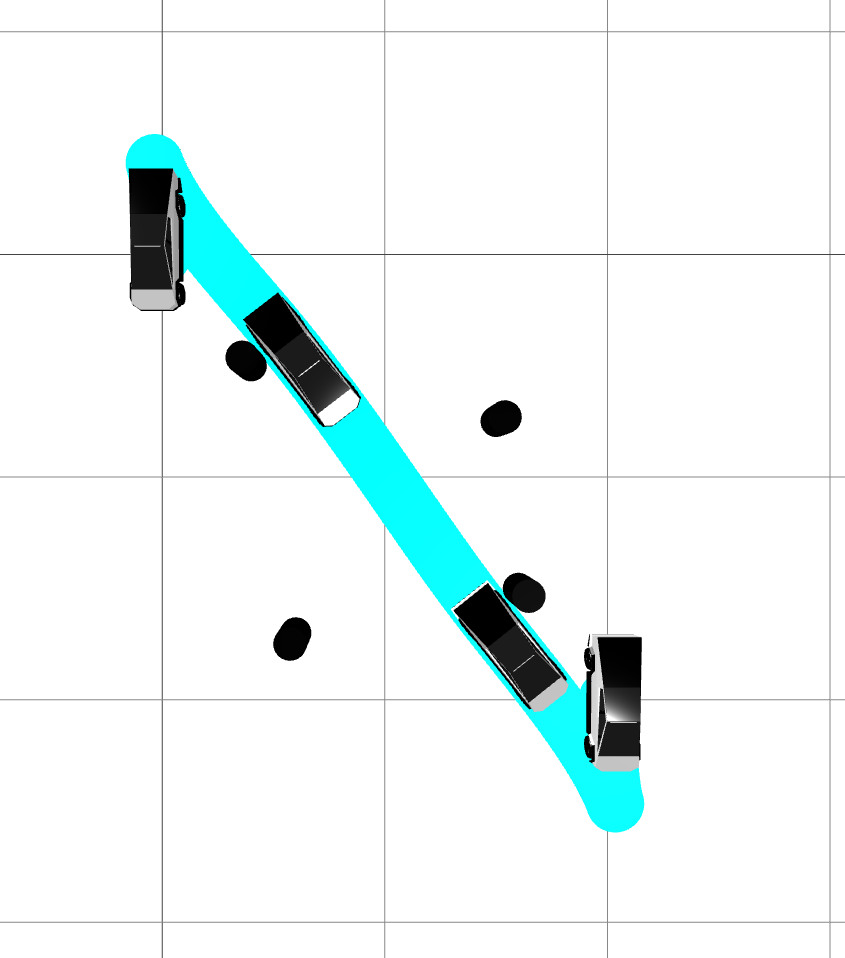}
		\hfill
		\includegraphics[width=0.1425\textwidth]{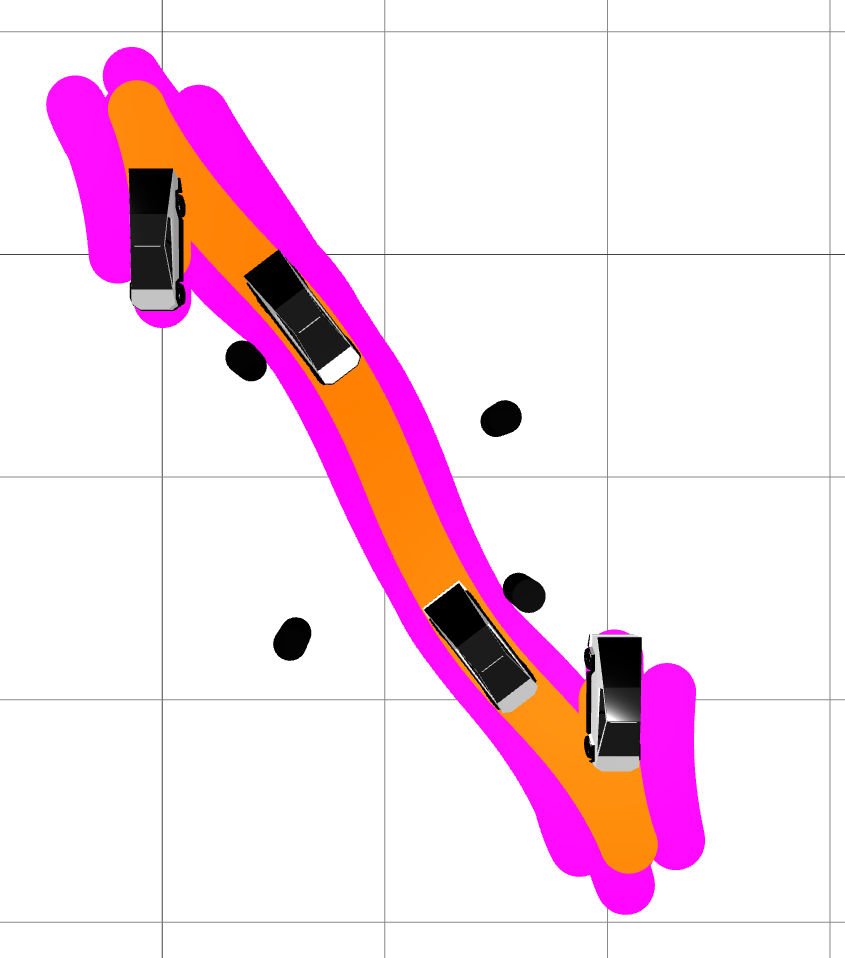}
		\hfill
		\includegraphics[width=0.1425\textwidth]{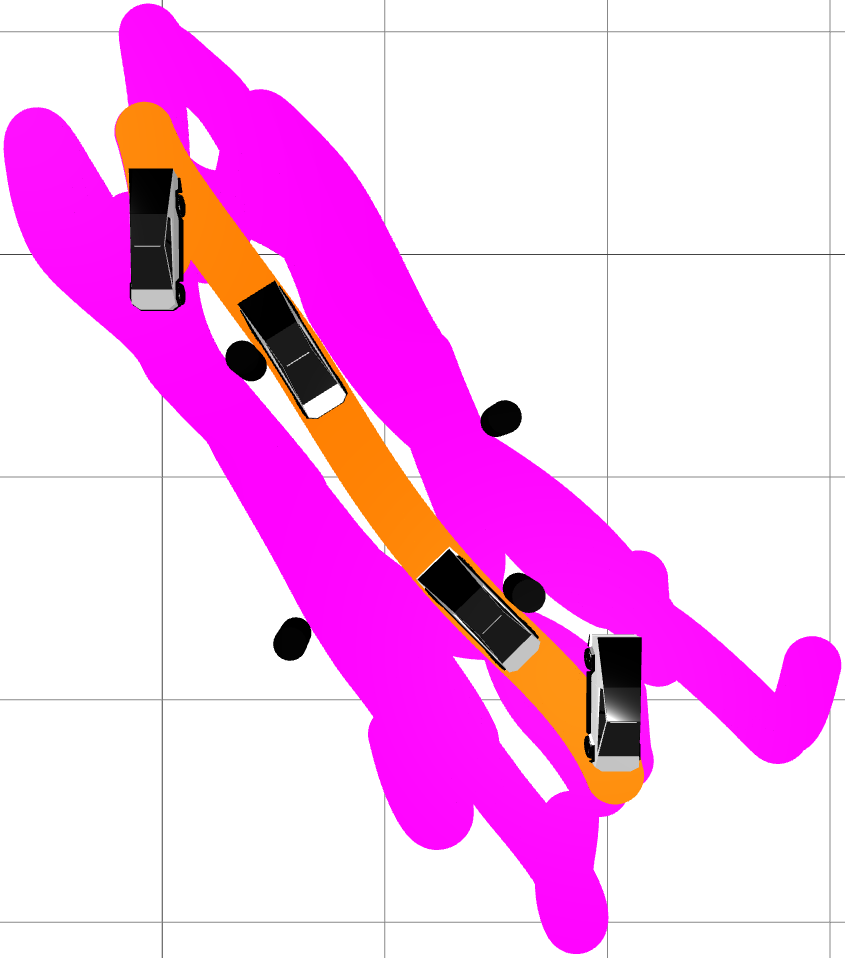}
		\caption{Position trajectories for autonomous car avoiding obstacles. TO (left) finds a path that is near the obstacles, whereas DPO with $\delta = 0.001$ (center) finds a path that avoids obstacles with a margin for safety. With larger disturbances, $\delta = 0.01$, samples (magenta) diverge and the DPO policy fails to safely avoid obstacles (right).}
		\label{car_trajectory}
	\end{figure}

\section{Direct Policy Optimization}\label{DPO}
	\rev{We now present the Direct Policy Optimization algorithm. DPO makes several strategic approximations to the discrete-time stochastic optimal-control problem: First, the expectation of the objective in \eqref{policy_opt} is approximated using the unscented transform. \rev{Second}, DPO explicitly optimizes a reference trajectory, $\bar{\tau}$, and $N$ sample trajectories, $\tau^{(i)}$, in order to approximate the stochastic dynamics. We use an overbar to denote reference or nominal quantities and superscripts to denote sample indices throughout the paper. Next, we seek a local feedback policy that is valid in the neighborhood of the reference trajectory. Finally, chance constraints are approximately enforced by applying inequality constraints to a set of sample points chosen from level sets of the state and input distributions}. Using these approximations, we formulate a nonlinear program (NLP) that is amenable to optimization with large-scale quasi-Newton solvers.
	
\subsection{\rev{Objective}}
	DPO minimizes the following \rev{objective}:
	\begin{equation} \label{dpo_cost}
		J(\bar{\tau}) + \mathbf{E} [ S(\bar{\tau}, \tau) ] ,
	\end{equation}
	with cost function $J$ applied to the \rev{reference} trajectory and a quadratic tracking-cost function:
	\begin{equation}
		\begin{aligned}
			S(\bar{\tau},\tau) =&~(x_T - \bar{x}_T)^T Q_T (x_T - \bar{x}_T) \\ 
			& + \sum \limits_{t=1}^{T-1} \{(x_t - \bar{x}_t)^T Q_t (x_t - \bar{x}_t) \\
			& + (u_t - \bar{u}_t)^T R_t (u_t - \bar{u}_t)\},
		\label{sample_cost}
		\end{aligned}
	\end{equation}
	\rev{$S : \mathbf{R}^z \times \mathbf{R}^z \rightarrow \mathbf{R}$}, with $Q_t \in \mathbf{S}_{+}^n$ and $R_t \in \mathbf{S}_{+}^m$, that penalizes deviations from the \rev{reference} trajectory under disturbances. After  simple manipulations, we can write \eqref{dpo_cost} in terms of the quantities introduced in Sec. \ref{ut}:
	\begin{multline}
		\mathbf{E} [ (x_t - \bar{x}_t)^T Q_t (x_t - \bar{x}_t) ] \\= \Tr(P_t Q_t) + (\mu_t - \bar{x}_t)^T Q_t (\mu_t - \bar{x}_t),
		\label{state_dev}
	\end{multline}
	\begin{multline}
		\mathbf{E} [ (u_t - \bar{u}_t)^T R_t (u_t - \bar{u}_t) ] \\= \Tr(L_t R_t) + (\nu_t - \bar{u}_t)^T R_t (\nu_t - \bar{u}_t).
		\label{control_dev}
	\end{multline}
	Since the expectation in \eqref{dpo_cost} depends only on the first and second moments of the state and control distributions, we can efficiently compute it using the unscented transform. As an aside, the second terms in \eqref{state_dev} and \eqref{control_dev} are zero in the linear-quadratic Gaussian (LQG) case, and their presence reflects the invalidity of the separation principle in more general settings.
	
	\begin{figure}[t]
		\centering
		\includegraphics{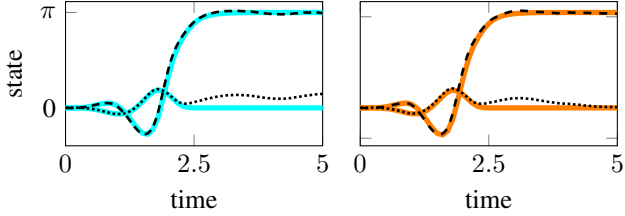}
		\caption{Simulated tracking (black) for position (dotted) and orientation (dashed) of cart-pole experiencing Coulomb friction. Comparison between LQR (blue) and DPO (orange) policies.}
		\label{cartpole_sim}
	\end{figure}

\subsection{Dynamics}
	\rev{DPO utilizes direct collocation and enforces dynamics for each trajectory via equality constraints at each time step. A unimodal distribution over the state and policy is maintained along the planning horizon by resampling the sample trajectories at each time step using the unscented transform (\ref{sigma_point_update}-\ref{sample_covariance_control}); this procedure is summarized in Algorithm \ref{unscented_dynamics}. For motion-planning applications, maintaining a unimodal distribution over a planning horizon is a reasonable modeling choice because the policy explicitly works to keep sample trajectories near the reference. In addition to computing terms required by the expectation over the objective, resampling at each time step prevents samples from collapsing to the reference at later time steps, encouraging the policy to be robust throughout the entire trajectory.}
		
	\rev{Initial sample states and disturbance inputs are deterministically drawn from normal distributions, $x_1^{(i)} \sim \mathcal{N}(\mu_1, P_1)$ and $w_t \sim \mathcal{N}(0, D_t)$, using (\ref{sigma_point_update}). The initial reference state is $\bar{x}_1 = \mu_1$. By utilizing the discrete-time dynamics, it is possible to generate non-Gaussian noise for the system and capture model uncertainty.}

\subsection{Feedback Policy}
	Each sample trajectory is subject to a policy constraint:
	\begin{equation}
		u_t^{(i)} = \pi_t(x_t^{(i)},\bar{x}_t,\bar{u}_t,\theta_t), \label{policy_constraint}
	\end{equation}
	at each time step. Instead of optimizing global policies, we search for local feedback policies that can depend on the reference trajectory. While these policies can be from any differentiable function class, we focus on linear policies for simplicity and leave extensions to more complex functions to future work.

\subsection{Chance Constraints}
	Chance constraints are approximated by constraining the \rev{reference} and sample trajectories. \rev{The probability of violation, $\epsilon$, corresponds to particular level sets of the state and control distributions. The sampling parameter $\beta$ can be selected in order to sample sigma points that approximate these level sets. Constraints are only enforced at these points. While more accurate methods for evaluating chance constraints exist, they are much more computationally demanding. Instead, DPO trades computational tractability for exact guarantees.} 
	
\subsection{Nonlinear Programming Formulation}
	DPO can be formulated as the following NLP: 
	\begin{alignat}{3}
	&\underset{\Theta, \bar{\tau}, \tau^{(1:N)}}{\mbox{minimize}} && \quad J(\bar{\tau}) + \sum \limits_{i=1}^{N} S(\bar{\tau}, \tau^{(i)}) \label{dpo}\\
	&\mbox{subject to} &&\quad \bar{x}_{t+1} = \bar{f}_t(\bar{x}_t,\bar{u}_t,0),&&\hspace{0.5em} t = 1,\dots,T-1, \notag\\
	&&&\quad x_{t+1}^{(1:N)} = g_t(\tau_t^{(1:N)}, D_t),&&\hspace{0.5em} t = 1,\dots,T-1,\notag \\
	&&&\quad u^{(1:N)}_t = \notag\\ &&&\quad\quad\pi_t(x^{(1:N)}_t,\bar{x}_t,\bar{u}_t,\theta_t),&&\hspace{0.5em} t = 1,\dots,T-1,\notag\\
	&&&\quad c_j(\bar{\tau}) \leq 0,&&\hspace{0.5em} j = 1,\dots,k,\notag\\
	&&&\quad c_j(\tau^{(1:N)}) \leq 0,&&\hspace{0.5em} j = 1,\dots,k.\notag
	\end{alignat}
	Additional constraints, for example, conditions on the policy parameters or trajectories, can be directly added to the formulation. Off-the-shelf large-scale NLP solvers like Ipopt~\cite{wachter2006implementation} and SNOPT~\cite{gill2005snopt} can be employed to efficiently optimize \eqref{dpo} by taking advantage of sparsity in its associated Jacobian and Hessian matrices.
	
	\begin{table}[t]
		\centering
		\caption{Tracking error, computed with (\ref{sample_cost}), for cart-pole experiencing Coulomb friction. Comparison between LQR and DPO policies.}
		\begin{tabular}{c c c}
			\toprule
			\textbf{cost} &
			\specialcellbold{LQR} &
			\specialcellbold{DPO} \\
			\toprule
			state & 3.18 & \textbf{2.26}\\
			control & 0.84 & \textbf{0.39}\\
			total & 4.02 & \textbf{2.64}\\
			\toprule
		\end{tabular}
		\label{cartpole_results}
	\end{table}

\section{Examples}\label{examples}
	The following examples were implemented in Julia, used SNOPT to optimize (\ref{dpo}), and were performed on a laptop computer with an Intel Core i7-7600U 2.80 GHz CPU and 16 GB of memory.
	
	To verify the performance of DPO, optimized policies are simulated at ten times the sample rate used during optimization, systems are simulated with explicit third-order Runge-Kutta integration, zero-order-hold control interpolation, cubic spline state interpolation, and are subject to additive zero-mean Gaussian noise. TO and DPO utilize the same objective for the reference trajectory. Throughout, the tracking cost (\ref{sample_cost}) and LQR policies have the same weights and, unless specified, $\beta = 1$. The reported tracking error is computed using the tracking cost (\ref{sample_cost}) for a single trajectory. During optimization we employ implicit midpoint integration and use discrete dynamics with additive noise:
	\begin{equation}
		x^{(i)}_{t+1} = f^{(i)}_t(x^{(i)}_t,u^{(i)}_t) + w^{(i)}_t.
		\label{dynamics_add_noise}
	\end{equation} We find that additive noise with state augmentation and different sample models is quite general and capable of capturing interesting dynamical effects. Policies use linear feedback to track the reference trajectory:
	\begin{equation}
		u^{(i)}_t = \bar{u}_t - \theta_t (x^{(i)}_t - \bar{x}_t). \label{linear_policy}
	\end{equation}
	Our implementation of DPO and additional details for each example are available at, \url{http://roboticexplorationlab.org/projects/dpo.html}.
		
	\begin{figure}[t]
		\centering
		\includegraphics[width=0.35\textwidth]{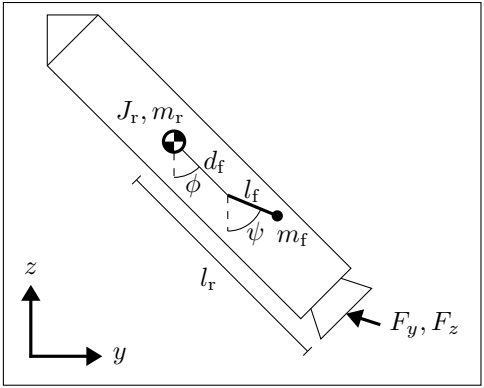}
		\caption{Rocket with planar dynamics. The model has lateral and vertical positions $y$ and $z$, orientation $\phi$, and thrust inputs $F_y$ and $F_z$. Parameters are inertia $J_{\mathrm{r}}$, mass $m_{\mathrm{r}}$, and length  $l_{\mathrm{r}}$ from center of mass to thruster. A pendulum positioned $d_{\mathrm{f}}$ from rocket center of mass, with orientation $\psi$, length $l_{\mathrm{f}}$, and mass $m_{\mathrm{f}}$, models fuel slosh.}
		\label{rocket_model}
	\end{figure}

\subsection{Double Integrator}
	We first demonstrate that DPO recovers the LQR solution for double integrator dynamics:
	\begin{equation}
		x_{t+1} = \begin{bmatrix} 1 & 1 \\ 0 & 1 \end{bmatrix}x_t + \begin{bmatrix} 0 \\ 1 \end{bmatrix}u_t + w_t, \label{double_integrator_dynamics}
	\end{equation}
	\rev{with $n = 2$ states, $m = 1$ controls, and $d = 2$ disturbances, regulated to the origin over a horizon $T = 51$. The policy has $p = 2$ parameters at each time step. Initial states are sampled from a distribution with $\mu_1 = 0$ and $P_1 = I$, disturbances have $D_{1:T-1} = I$, the weights are $Q_{1:T} = I$ and $R_{1:T-1} = I$.}
	
	The decision variables are initialized by uniformly sampling between $-1$ and $1$ for 1000 trials. The maximum, mean, and standard deviations of the normalized error between the DPO policies and the exact LQR solution computed with the Riccati equation are near the tolerance of the solver: 2.4e-5, 4.0e-7, 8.5e-7, respectively---reinforcing that the approximations made in the derivation of DPO are exact in the LQG case. Additionally, the second term in \eqref{state_dev} is zero, demonstrating that the separation principle holds for this problem.

\subsection{Car}
	We plan a collision-free path through four obstacles for an autonomous car, modeled as a unicycle~\cite{lavalle2006planning}:
	\begin{equation}
		\begin{bmatrix} \dot{y} \\ \dot{z} \\ \dot{\phi} \end{bmatrix} = \begin{bmatrix} v \text{cos}(\phi) \\ v \text{sin}(\phi) \\ \omega \end{bmatrix}. \label{unicycle_dynamics}
	\end{equation}
	The $n = 3$ states are positions $y$ and $z$, and orientation $\phi$. The $m = 2$ controls are the forward and angular velocities, $v$ and $\omega$. There are $d = 3$ disturbance inputs and the policy has $p = 6$ parameters at each time step. Initial states are sampled from a distribution with $\mu_1 = 0$, $P_1 = \textbf{diag}(1,1,0.1)$, weights are $Q_{1:T-1} = \textbf{diag}(10,10,1)$, $Q_T = 100I$, and $R_{1:T-1} = 0.1I$, and disturbances have $D_{1:T-1} = \textbf{diag}(\delta, \delta, 0.1\delta)$. The planning horizon is $T = 51$ with fixed time step $h = 0.02$. The choice of $\beta = 1$ over the planning horizon results in constraints being enforced on the 1-sigma level set of the state distribution, corresponding to $\epsilon = 0.32$.
	
	For this scenario, TO finds a path that is in close proximity to the obstacles. In contrast, by considering noise, DPO with $\delta = 0.001$ finds a path that remains a safe distance from the obstacles. Decreasing the noise results in convergence to the TO solution. However, larger disturbance inputs (e.g., $\delta = 0.01$), expose a limitation of DPO;  samples can diverge, resulting in a multimodal state distribution with trajectories taking different paths around the obstacles (Fig. \ref{car_trajectory}). Ultimately, unlike \textit{ad hoc} techniques like obstacle inflation, DPO considers the coupling of dynamics, constraints, and disturbances to generate safer paths for the system.
	
	\begin{figure}[t]
		\centering
		\includegraphics{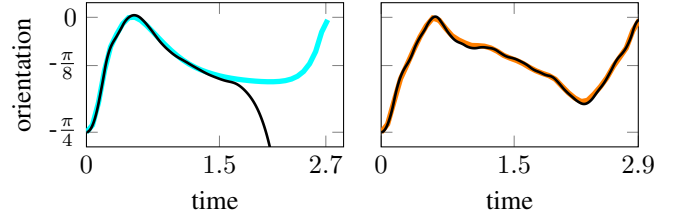}
		\caption{Simulated tracking (black) for the orientation of a rocket experiencing fuel slosh during landing. Comparison between LQR (blue) and DPO (orange) polices.}
		\label{rocket_tracking}
	\end{figure}
	
\subsection{Cart-Pole}
	A swing-up trajectory for a cart-pole~\cite{tedrake2014underactuated} with a slider that experiences Coulomb friction is synthesized for horizon $T = 51$ and fixed time step $h = 0.1$. The state, $x = (y, \phi, \dot{y}, \dot{\phi})$, has cart position $y$, pendulum orientation $\phi$, and their respective time derivatives. The optimality conditions for the maximum-dissipation principle~\cite{moreau2011unilateral} are used as constraints to explicitly model friction~\cite{manchester2020variational} with coefficient $\mu_{\text{f}} = 0.1$ for each trajectory. \rev{The cart position is controlled and $m = 1$,  there are $d = 4$ disturbance inputs, and the policy has $p = 4$ parameters at each time step. Initial states are sampled from $\mu_1 = 0$ and $P_1 = I$, the weights are $Q_{1:T-1} = \textbf{diag}(10,10,1,1)$, $Q_T = 100I$, and $R_{1:T-1}= I$, and disturbances have $D_{1:T-1} = 0.001 I$.}

	The performance of LQR tracking \rev{reference} trajectories optimized subject to friction, as well as the DPO policy, are verified in simulation. Results are provided in Table \ref{cartpole_results} and tracking for the position and orientation is shown in Fig. \ref{cartpole_sim}. By first designing a trajectory that explicitly models friction, it is possible to subsequently synthesize an LQR policy for tracking. In contrast, DPO, which can handle nonsmooth dynamics, produces superior tracking by simultaneously optimizing the \rev{reference} trajectory and policy.
	
	\begin{table}[t]
		\centering
		\caption{Tracking error, computed with (\ref{sample_cost}), for rocket landing with fuel slosh. Comparison between LQR and DPO policies.}
		\begin{tabular}{c c c}
			\toprule
			\textbf{cost} &
			\specialcellbold{LQR} & 
			\specialcellbold{DPO} \\
			\toprule
			state & 4725.39 & \textbf{5.02}\\
			control & 75.21 & \textbf{0.39}\\
			total & 4800.60 & \textbf{5.41}\\
			\toprule
		\end{tabular}
		\label{rocket_results}
		\vspace{-0.5cm}
	\end{table}

\subsection{Rocket Landing}
	We plan a soft landing for a rocket~\cite{meditch1964problem}. The nominal system with planar dynamics has state $x = (y,z,\phi,\dot{y},\dot{z},\dot{\phi})$, with lateral and vertical positions $y$ and $z$, orientation $\phi$, and their respective time derivatives. The rocket is controlled with gimballed thrust, $u = (F_{y},F_{z})$. It is initialized with non-zero displacements and velocities relative to its target state: a zero-velocity, vertical-orientation touchdown inside a landing zone.	
	
	The rocket experiences fuel slosh during landing that is not accounted for in the nominal model. This is a critical dynamical effect, but it is difficult to model and observe. In practice, these unmodeled effects are typically handled in \textit{ad hoc} ways, often with extensive Monte Carlo simulation and controller tuning. To approximately model fuel slosh, we augment the nominal model with two additional states associated with an unobservable and unactuated pendulum (Fig. \ref{rocket_model}). A common frequency-domain control approach is to include a notch filter at the pendulum's natural frequency in order to prevent excitation of these fuel-slosh dynamics. We instead use DPO to synthesize a policy that is robust to fuel slosh.

	\rev{The \rev{reference} trajectory is optimized with the nominal model, while the sample trajectories are subject to the augmented model in order to capture fuel-slosh dynamics, and a linear output feedback policy is optimized that does not have access to the fuel-slosh states. There are $d = 8$ disturbance inputs to the augmented model and the policy has $p = 12$ parameters at each time step. Initial states for the augmented model are sampled with uncertainty $P_1 = I$, weights are $Q_{1:T-1} = 100I$, $Q_T = 1000I$, and $R_{1:T-1} = \textbf{diag}(1,1,100)$, and disturbances have $D_{1:T-1} = 0.001I$. Over the planning horizon $T = 41$ with free final time, thrust limits are enforced.}
	
	TO finds a 2.72 second solution, whereas DPO finds a longer 2.91 second path to the landing zone. The position trajectories are shown in Fig. \ref{rocket_trajectory}, simulation results with fuel slosh are provided in Table \ref{rocket_results}, and tracking results for the rocket's orientation are shown in Fig. \ref{rocket_tracking}. Due to fuel slosh, LQR fails to track the path generated by TO and the rocket tips over, whereas the DPO policy successfully lands the rocket. DPO is able to optimize a policy for a system with unobservable dynamical effects by sampling augmented models.
	
	\begin{figure}[t]
		\centering
		\includegraphics[width=.45\textwidth]{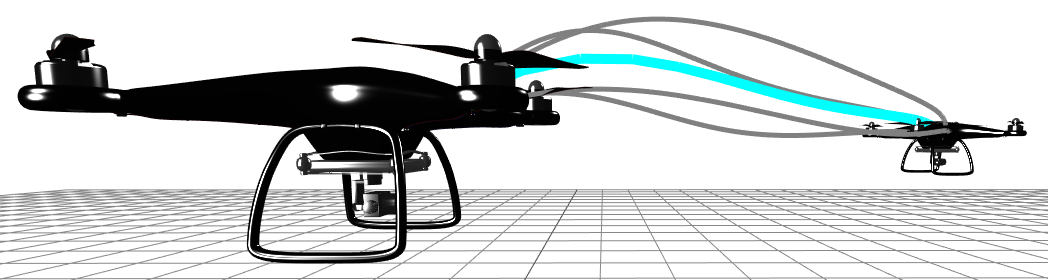}
		
		\vspace{0.01\textwidth}
		
		\includegraphics[width=.45\textwidth]{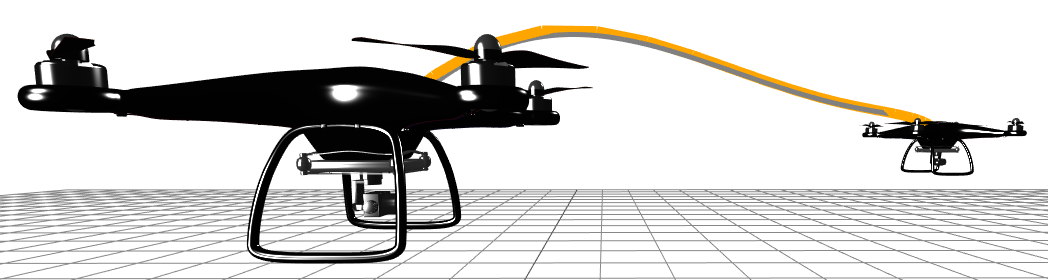}
		
		\caption{Simulated position trajectories (gray) for quadrotor controlled with LQR (top) and DPO (bottom) policies while experiencing different propellers breaking. LQR (blue) has degrading tracking while DPO (orange) has superior, and nearly identical tracking regardless of which propeller breaks.}
		\label{quadrotor_prop}
	\end{figure}

\subsection{Quadrotor}
	A point-to-point maneuver is planned for a quadrotor~\cite{mellinger2012trajectory} that experiences a random blade breaking off from a propeller during flight. A broken propeller is modeled by constraining the corresponding control input to be half its nominal maximum value. We use DPO to optimize a policy that is robust to this event occurring for any of the propellers. The reference model has no broken propellers, but the sample trajectories are optimized with different models from four groups, each experiencing a different broken propeller. \rev{There are $n = 12$ states, $m = 4$ controls, $d = 12$ input disturbances, and the policy has $p = 48$ parameters at each time step. Initial states are sampled around the initial nominal state with uncertainty $P_1 = I$, weights are $Q_{1:T-1} = 10I$, $Q_T = 100I$, and $R_{1:T-1} = I$, and disturbances have $D_{1:T-1} = 0.001 I$. Thrust limits are enforced over the planning horizon $T=31$ with a free final time.}

\begin{figure}[t]
		\centering
		\includegraphics{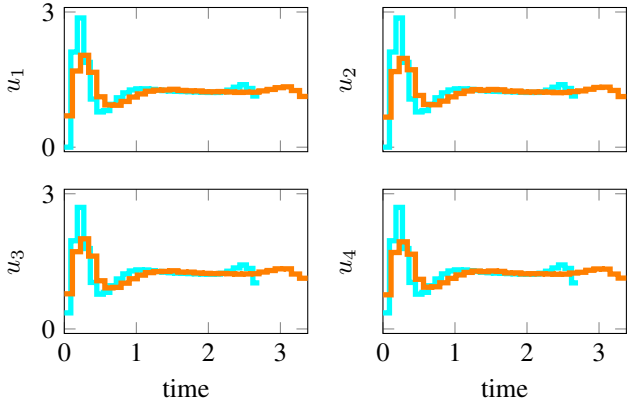}
		\caption{Nominal controls for quadrotor performing point-to-point maneuver generated with TO (blue) and DPO (orange). The controls found with DPO are applied over a longer horizon and have lower maximum values compared to TO.}
		\label{quadrotor_control}
	\end{figure}

	TO finds a 2.71 second trajectory, while DPO finds a longer 3.38 second trajectory with lower maximum controls (Fig. \ref{quadrotor_control}). The policies are compared in simulation over 100 initial conditions with noise sampled from $\mathcal{N}(0, 0.001 I)$. One set of initial conditions is visualized in Fig. \ref{quadrotor_prop}. The average tracking performance over each propeller breaking is provided in Table \ref{quad_prop_results}. DPO finds a policy with consistent and superior tracking compared with LQR for all cases by optimizing over different models.
	
\section{Discussion} \label{discussion}
	We empirically demonstrate through Monte Carlo simulations that DPO recovers LQR policies for a double integrator system. It seems possible that for LQG problems, despite the non-convexity of the policy constraints, there exists a unique global solution for DPO. However, we leave a formal convergence analysis to future work.

	The motion-planning examples highlight a number of DPO's capabilities compared to DIRTREL and GPS. In the cart-pole example, DPO is able to subject each trajectory to discontinuous Coulomb friction calculated with the maximum-dissipation principle, whereas a similar DIRTREL example only applied white-noise disturbances to a single model~\cite{manchester2019robust}. This example also highlights how jointly optimizing trajectories and a policy leads to superior performance, a distinction compared to the decoupled approach employed by GPS. Next, in the rocket landing example, DPO optimizes an output feedback policy, whereas DIRTREL is limited to state feedback. Additionally, directly modeling fuel-slosh dynamics with a pendulum is likely simpler than designing input disturbances for DIRTREL to capture the same effects. Finally, with DPO, additional constraints can be applied to any of the decision variables, whereas the ability of GPS to handle constraints using DDP and stochastic gradient descent is limited.
	
	Because we use local optimization methods to solve DPO, providing good initial guesses to the solver is crucial for performance of the algorithm. In practice, we use a standard TO solution to warm start the \rev{reference} and sample trajectories for DPO. For linear state-feedback policies, we use the corresponding LQR solution as an initial guess. While not necessary in any of our examples, initial guesses for more complex policies could be found by running DPO without the policy constraint, then performing an offline regression to fit approximate policy parameters for warm starting. 
	
	For a TO problem solved with a sparsity-exploiting second-order method, the computational complexity is approximately $O\Big( T (n + m)^3 \Big)$~\cite{wang2009fast}. For DPO, we can consider an augmented state and control with dimensions $n(2n + 2)$ and $m(2n + 2)$, respectively. The complexity of DPO is, therefore, $O\Big( T (n^6 + m^3n^3) \Big)$. In the examples, which do not employ a specialized solver, solution times range from seconds to 1.5 hours for a state with $n = 12$ on a laptop computer. It is likely possible to exploit the DPO problem's structure to improve the complexity \rev{by using custom linear solvers}. First-order methods, which can scale much better, become attractive for DPO as problems become large.
	
	Lastly, LQR is a powerful tool and can likely be tuned to qualitatively match the performance of the linear policies found with DPO in many cases. However, the strength of DPO lies in its ability to explicitly reason about robustness and the complex coupling between dynamics, constraints, and disturbances during synthesis, instead of relying on hand tuning and heuristics.
	
\section{Conclusion}\label{conclusion}
	\begin{table}[t]
		\centering
		\caption{\rev{Tracking errors with mean ($\mu$) and standard deviation ($\sigma$), computed with (\ref{sample_cost}), comparing LQR and DPO policies for quadrotor over 100 initial conditions for each propeller being broken.}}
		\begin{subtable}{0.5\textwidth}
			\centering
			\begin{tabular}{c c c}
				\toprule
				\textbf{cost ($\mu$, $\sigma$)} &
				\specialcellbold{LQR} & 
				\specialcellbold{DPO} \\
				\toprule
				state & (13.43, 12.64) & \textbf{(4.74, 5.87)} \\
				control & (0.074, 0.067) & \textbf{(0.020, 0.022)} \\
				total & (13.50, 12.71) & \textbf{(4.76, 5.89)}\\
				\toprule
			\end{tabular}
			\caption{propeller 1 broken}
		\end{subtable}
		\vfill
		\begin{subtable}{0.5\textwidth}
			\centering
			\begin{tabular}{c c c}
				\toprule
				\textbf{cost ($\mu$, $\sigma$)} &
				\specialcellbold{LQR} & 
				\specialcellbold{DPO} \\
				\toprule
				state & (13.13, 11.98) & \textbf{(4.74, 5.87)} \\
				control & (0.073, 0.065) & \textbf{(0.020, 0.023)} \\
				total & (13.21, 12.05) & \textbf{(4.76, 5.89)}\\
				\toprule
			\end{tabular}
			\caption{propeller 2 broken}
		\end{subtable}
		\vfill
		\begin{subtable}{0.5\textwidth}
			\centering
			\begin{tabular}{c c c}
				\toprule
				\textbf{cost ($\mu$, $\sigma$)} &
				\specialcellbold{LQR} & 
				\specialcellbold{DPO} \\
				\toprule
				state & (3431.91, 12895.93) & \textbf{(4.74, 5.88)} \\
				control & (2.21, 7.86) & \textbf{(0.020, 0.023)} \\
				total & (3434.12, 12903.75) & \textbf{(4.76, 5.90)}\\
				\toprule
			\end{tabular}
			\caption{propeller 3 broken}
		\end{subtable}
		\vfill
		\begin{subtable}{0.5\textwidth}
			\centering
			\begin{tabular}{c c c}
				\toprule
				\textbf{cost ($\mu$, $\sigma$)} &
				\specialcellbold{LQR} & 
				\specialcellbold{DPO} \\
				\toprule
				state & (4843.89, 14892.19) & \textbf{(4.74, 5.87)} \\
				control & (3.21, 9.41) & \textbf{(0.020, 0.023)} \\
				total & (4847.10, 14901.59) & \textbf{(4.76, 5.89)}\\
				\toprule
			\end{tabular}
			\caption{propeller 4 broken}
		\end{subtable}
		\label{quad_prop_results}
	\end{table}
	We have presented a new algorithm, Direct Policy Optimization, for approximately solving stochastic optimal-control problems. \rev{We demonstrate that the} algorithm is exact in the LQG case and is capable of optimizing policies for nonlinear systems that respect control limits and obstacles, and are robust to unmodeled dynamics.
	
	Many interesting avenues for future work remain: Extensions to nonlinear policies could be made by introducing high-dimensional features and regularization or constraints on policy parameters, or neural networks could be used to parameterize policies since the policy parameters scale much better compared to state and control dimensions. Another potential direction is to optimize a local value function approximation in place of an explicit policy. A much richer class of disturbances and model parameter uncertainty could also be modeled by augmenting the state vector and dynamics. Lastly, more complex systems with larger state and input dimensions may be more amenable to optimization with first-order or matrix-free methods.

%\AtNextBibliography%{\footnotesize}
%\printbibliography
\bibliographystyle{IEEEtran}  % can change to fit your field e.g. pnas2009
\bibliography{main} % no .bib extension necessary

\begin{thebibliography}{10}
\providecommand{\url}[1]{#1}
\csname url@samestyle\endcsname
\providecommand{\newblock}{\relax}
\providecommand{\bibinfo}[2]{#2}
\providecommand{\BIBentrySTDinterwordspacing}{\spaceskip=0pt\relax}
\providecommand{\BIBentryALTinterwordstretchfactor}{4}
\providecommand{\BIBentryALTinterwordspacing}{\spaceskip=\fontdimen2\font plus
\BIBentryALTinterwordstretchfactor\fontdimen3\font minus
  \fontdimen4\font\relax}
\providecommand{\BIBforeignlanguage}[2]{{%
\expandafter\ifx\csname l@#1\endcsname\relax
\typeout{** WARNING: IEEEtran.bst: No hyphenation pattern has been}%
\typeout{** loaded for the language `#1'. Using the pattern for}%
\typeout{** the default language instead.}%
\else
\language=\csname l@#1\endcsname
\fi
#2}}
\providecommand{\BIBdecl}{\relax}
\BIBdecl

\bibitem{bertsekas2004stochastic}
D.~P. Bertsekas and S.~Shreve, \emph{{{S}tochastic Optimal Control: {T}he
  Discrete-Time Case}}.\hskip 1em plus 0.5em minus 0.4em\relax Athena
  Scientific, 2004.

\bibitem{stryk1993numerical}
O.~Von~Stryk, ``Numerical solution of optimal control problems by direct
  collocation,'' in \emph{Optimal Control}.\hskip 1em plus 0.5em minus
  0.4em\relax Springer, 1993, pp. 129--143.

\bibitem{jacobson1970differential}
D.~H. Jacobson and D.~Q. Mayne, \emph{Differential Dynamic Programming}.\hskip
  1em plus 0.5em minus 0.4em\relax Elsevier Publishing Company, 1970, no.~24.

\bibitem{kuindersma2016optimization}
S.~Kuindersma, R.~Deits, M.~Fallon, A.~Valenzuela, H.~Dai, F.~Permenter,
  T.~Koolen, P.~Marion, and R.~Tedrake, ``Optimization-based locomotion
  planning, estimation, and control design for the {A}tlas humanoid robot,''
  \emph{Autonomous Robots}, vol.~40, no.~3, pp. 429--455, 2016.

\bibitem{moore2014robust}
J.~L. Moore, ``Robust post-stall perching with a fixed-wing {UAV},'' Ph.D.
  dissertation, Massachusetts Institute of Technology, 2014.

\bibitem{manchester2019robust}
Z.~Manchester and S.~Kuindersma, ``Robust direct trajectory optimization using
  approximate invariant funnels,'' \emph{Autonomous Robots}, vol.~43, no.~2,
  pp. 375--387, 2019.

\bibitem{todorov2005generalized}
E.~Todorov and W.~Li, ``A generalized iterative {LQG} method for
  locally-optimal feedback control of constrained nonlinear stochastic
  systems,'' in \emph{American Control Conference}, 2005, pp. 300--306.

\bibitem{silver2014deterministic}
D.~Silver, G.~Lever, N.~Heess, T.~Degris, D.~Wierstra, and M.~Riedmiller,
  ``Deterministic policy gradient algorithms,'' in \emph{International
  Conference on Machine Learning}, 2014.

\bibitem{williams1992simple}
R.~J. Williams, ``Simple statistical gradient-following algorithms for
  connectionist reinforcement learning,'' \emph{Machine Learning}, vol.~8, no.
  3-4, pp. 229--256, 1992.

\bibitem{tobin2017domain}
J.~Tobin, R.~Fong, A.~Ray, J.~Schneider, W.~Zaremba, and P.~Abbeel, ``Domain
  randomization for transferring deep neural networks from simulation to the
  real world,'' in \emph{IEEE/RSJ International Conference on Intelligent
  Robots and Systems}, 2017, pp. 23--30.

\bibitem{mania2018simple}
H.~Mania, A.~Guy, and B.~Recht, ``Simple random search of static linear
  policies is competitive for reinforcement learning,'' in \emph{Advances in
  Neural Information Processing Systems}, 2018, pp. 1800--1809.

\bibitem{bu2019lqr}
J.~Bu, A.~Mesbahi, M.~Fazel, and M.~Mesbahi, ``{LQR} through the lens of first
  order methods: {Discrete}-time case,'' \emph{arXiv:1907.08921}, 2019.

\bibitem{wytock2013fast}
M.~Wytock and J.~Z. Kolter, ``A fast algorithm for sparse controller design,''
  \emph{arXiv:1312.4892}, 2013.

\bibitem{agrawal2020learning}
A.~Agrawal, S.~Barratt, S.~Boyd, and B.~Stellato, ``Learning convex
  optimization control policies,'' in \emph{Learning for Dynamics and Control},
  2020, pp. 361--373.

\bibitem{amos2018differentiable}
B.~Amos, I.~Jimenez, J.~Sacks, B.~Boots, and J.~Z. Kolter, ``Differentiable
  {MPC} for end-to-end planning and control,'' in \emph{Advances in Neural
  Information Processing Systems}, 2018, pp. 8289--8300.

\bibitem{levine2013guided}
S.~Levine and V.~Koltun, ``Guided policy search,'' in \emph{International
  Conference on Machine Learning}, 2013, pp. 1--9.

\bibitem{uhlmann1995dynamic}
J.~K. Uhlmann, ``Dynamic map building and localization: {N}ew theoretical
  foundations,'' Ph.D. dissertation, University of Oxford, 1995.

\bibitem{menegaz2015systematization}
H.~M.~T. Menegaz, J.~Y. Ishihara, G.~A. Borges, and A.~N. Vargas, ``A
  systematization of the unscented {K}alman filter theory,'' \emph{IEEE
  Transactions on Automatic Control}, vol.~60, no.~10, pp. 2583--2598, 2015.

\bibitem{wachter2006implementation}
A.~W{\"a}chter and L.~T. Biegler, ``On the implementation of an interior-point
  filter line-search algorithm for large-scale nonlinear programming,''
  \emph{Mathematical Programming}, vol. 106, no.~1, pp. 25--57, 2006.

\bibitem{gill2005snopt}
P.~E. Gill, W.~Murray, and M.~A. Saunders, ``{SNOPT}: {A}n {SQP} algorithm for
  large-scale constrained optimization,'' \emph{SIAM Review}, vol.~47, no.~1,
  pp. 99--131, 2005.

\bibitem{lavalle2006planning}
S.~M. LaValle, \emph{Planning Algorithms}.\hskip 1em plus 0.5em minus
  0.4em\relax Cambridge University Press, 2006.

\bibitem{tedrake2014underactuated}
R.~Tedrake, ``Underactuated robotics: {A}lgorithms for walking, running,
  swimming, flying, and manipulation (course notes for {MIT} 6.832),'' 2022.

\bibitem{moreau2011unilateral}
J.~J. Moreau, ``On unilateral constraints, friction and plasticity,'' in
  \emph{New Variational Techniques in Mathematical Physics}.\hskip 1em plus
  0.5em minus 0.4em\relax Springer, 2011, pp. 171--322.

\bibitem{manchester2020variational}
Z.~Manchester and S.~Kuindersma, ``Variational contact-implicit trajectory
  optimization,'' in \emph{Robotics Research}.\hskip 1em plus 0.5em minus
  0.4em\relax Springer, 2020, pp. 985--1000.

\bibitem{meditch1964problem}
J.~S. Meditch, ``On the problem of optimal thrust programming for a lunar soft
  landing,'' \emph{IEEE Transactions on Automatic Control}, vol.~9, no.~4, pp.
  477--484, 1964.

\bibitem{mellinger2012trajectory}
D.~Mellinger, N.~Michael, and V.~Kumar, ``Trajectory generation and control for
  precise aggressive maneuvers with quadrotors,'' \emph{The International
  Journal of Robotics Research}, vol.~31, no.~5, pp. 664--674, 2012.

\bibitem{wang2009fast}
Y.~Wang and S.~Boyd, ``Fast model predictive control using online
  optimization,'' \emph{IEEE Transactions on Control Systems Technology},
  vol.~18, no.~2, pp. 267--278, 2009.

\end{thebibliography}

\end{document}